\pdfoutput=1
\documentclass[10pt]{article}

\usepackage[utf8]{inputenc}
\usepackage[T1]{fontenc}
\usepackage[margin=1in]{geometry}
\usepackage{times}
\usepackage[round]{natbib}
\usepackage{hyperref}
\usepackage{url}
\usepackage{booktabs}
\usepackage{amsfonts}
\usepackage{amsmath}
\usepackage{amssymb}
\usepackage{graphicx}
\usepackage{subcaption}
\usepackage{microtype}
\usepackage{xcolor}
\usepackage{xspace}
\usepackage[percent]{overpic}
\usepackage{listings}
\usepackage{cleveref}
\usepackage{microtype}
\usepackage{lettrine}

\definecolor{primaryblue}{RGB}{0,77,153}
\definecolor{accentorange}{RGB}{198,89,17}

\usepackage[most]{tcolorbox}
\definecolor{softgray}{RGB}{128,128,128}
\definecolor{codebg}{RGB}{248,248,248}
\definecolor{successgreen}{RGB}{56,118,29}
\newtcolorbox{promptbox}[1][]{
    enhanced,
    colback=codebg,
    colframe=softgray,
    fonttitle=\bfseries\small\ttfamily,
    title={#1},
    boxrule=0.5pt,
    arc=2pt,
    left=6pt, right=6pt, top=4pt, bottom=4pt
}

\lstdefinelanguage{json}{
    backgroundcolor=\color{codebg},
    basicstyle=\small\ttfamily,
    breakatwhitespace=false,
    breaklines=true,
    captionpos=t,
    commentstyle=\color{successgreen},
    comment=[l]{//},
    frame=single,
    rulecolor=\color{softgray},
    keepspaces=true,
    keywordstyle=\color{primaryblue}\bfseries,
    numbers=left,
    numbersep=5pt,
    numberstyle=\tiny\color{softgray},
    showspaces=false,
    showstringspaces=false,
    showtabs=false,
    stringstyle=\color{accentorange},
    tabsize=2,
    xleftmargin=10pt,
    framexleftmargin=10pt
}

\newcommand{\methodname}{JODA\xspace}

\title{\methodname: Composable Joint Dynamics for Articulated Objects}

\author{%
  Tianhong Gao\textsuperscript{1}, Cheng Yu\textsuperscript{1}, Yinghao Xu\textsuperscript{2}, Mengyu Chu\textsuperscript{1}\\
  \textsuperscript{1}Peking University\\
  \textsuperscript{2}Ant Group, Robbyant
}

\date{}

\begin{document}

\maketitle

\begin{abstract}
Articulated objects used in simulation and embodied AI are typically
specified by geometry and kinematic structure,
but lack the fine-grained dynamical effects that govern realistic mechanical
behavior, such as frictional holding, detents, soft closing, and snap latching.
Existing approaches either ignore the detailed structure of dynamics
entirely, or use simple models with limited expressiveness.

We introduce JODA, a framework for generating joint-level dynamics as a structured three-channel field over the joint degree of freedom, capturing conservative forces, dry friction, and damping. Instantiated using shape-constrained piecewise cubic interpolation (PCHIP), this formulation defines a compact and expressive function space that is both interpretable and compatible with differentiable simulation. Building on this representation, we develop methods for inferring and refining joint dynamics from multimodal inputs. Given visual observations and joint context, a vision-language model proposes structured dynamical primitives, which are composed into a unified dynamics field. The resulting representation supports both direct manipulation and gradient-based refinement.
We demonstrate that JODA enables plausible and controllable modeling of diverse joint behaviors, providing a unified interface for inference, editing, and optimization. Code and example assets with their generated profiles will be released upon publication.
\end{abstract}

\section{Introduction}

Articulated-body assets are a central substrate for physical simulation,
robotics, embodied AI, and interactive scene generation. Recent asset pipelines~\citep{mandi2024real2code,he2026spark}
can recover or synthesize increasingly rich visual geometry and kinematic
structure: parts are separated, joints are assigned axes and ranges, and assets
can often be loaded by a physics engine. 
However, real-world interaction of articulated objects depends on more than kinematics alone. A
laptop lid holds at many intermediate angles; a microwave door may pop through
a closing latch; a refrigerator door can be pulled into its final closed pose;
an oven door may return under a broad spring-like tendency while being resisted
by friction. 
These behaviors reflect not only how objects are geometrically connected, but more importantly, how they behave dynamically.
Such dynamic effects are typically absent from automatic pipelines and must instead be manually authored or carefully tuned at significant cost, making them difficult to scale across datasets, scenes, and applications. As a result, automatic pipelines remain largely restricted to geometry and kinematic structure, leaving physical dynamics poorly modeled and leading to a persistent sim-to-real gap.

This paper studies the problem of generating \emph{dynamical structures} for
articulated-body assets: the structured, joint-level
internal effects that govern motion beyond rigid-body inertia, gravity, and hard joint limits. 
Rather than recovering the exact hidden mechanical assemblies inside an
object (e.g., specific springs or gears), we instead model their equivalent joint-level effects. This
distinction is crucial: mechanisms such as magnetic catches or spring latches may not be explicitly encoded by the asset geometry, yet their behavior can be expressed as position-dependent forces,
friction, or damping over the joint degree of freedom. 
This unified representation abstracts away low-level mechanical complexity while preserving macroscopic interactive fidelity.

A central question is how such structured dynamics can be initialized without direct supervision or manual design. Interestingly, we find that vision-language models (VLMs) capture useful implicit physical priors about object behavior; for instance, a VLM can infer an oven's tendency toward a stable closed configuration. However, these priors are latent and unstructured.
Our key insight is that the challenge is not in {eliciting} physical intuition from VLMs, but in {grounding} it within a structured, composable representation suitable for simulation.

We introduce \methodname, a framework that combines a composable dynamics representation with multimodal inference and refinement. Given an articulated asset and visual observations across joint configurations, a VLM proposes joint-level dynamical effects as structured hypotheses, including effect types, active intervals, and qualitative strengths. These hypotheses are then composed into a unified dynamics field. The resulting representation is compact, interpretable, and compatible with physical simulation, and supports both semantic and continuous refinement, including direct editing and gradient-based optimization.
Our approach unifies representation, inference, and optimization within a
single framework:
\begin{itemize}
    \item \textbf{Structured Joint Representation.} A composable representation for joint dynamics as a structured function space over the joint degree of freedom, enabling expressive yet interpretable modeling of forces, friction, and damping.
  \item \textbf{Multimodal Inference.} A pipeline that leverages VLMs to instantiate and iteratively refine joint dynamics, enabling reliable and diverse modeling.
  \item \textbf{Unified Interaction and Optimization.} \methodname supports flexible interaction and refinement, including direct or semantic editing and gradient-based optimization via differentiable simulation, providing a unified interface for inference, editing, and optimization.
\end{itemize}

\section{Related work}

\paragraph{Articulated object and asset generation}
Articulated object research has progressed from mobility prediction and
part-level datasets to automatic URDF-like reconstruction and generation from
images, videos, or language
\citep{hu2017mobility,xiang2020sapien,chen2024urdformer,liu2024singapo,mandi2024real2code, le2025articulateanything, xia2025drawerdigitalreconstructionarticulation, he2026spark}.
These methods primarily recover the geometric and kinematic scaffold of an
asset: parts, hierarchy, joint type, axis, origin, and motion limits.
\methodname is complementary: it assumes this scaffold is available and
generates the missing joint-level dynamical structure that determines how the
articulation behaves during simulation.

\paragraph{Simulation-ready assets and differentiable physics}
Physics simulators provide the substrate for executable articulated objects
\citep{xiang2020sapien,todorov2012mujoco,makoviychuk2021isaacgym,deepmind2024mjx,freeman2021brax,howell2022dojo}.
They expose primitives such as PD-drive controllers, tendons, and runtime
user-specified generalized forces. The built-in mechanisms are well suited to
motors or tendon actuation, but are not a convenient way to express everyday
object dynamics such as detents, endpoint attraction, or soft-close damping;
using external forces still requires hand-designed force laws.

Asset-centric benchmarks apply such simulators to large collections of
interactive objects and tasks \citep{gu2023maniskill2,li2024behavior1k}.
Simulation-trained robot policies instantiate the same pattern
\citep{zakka2023robopianist,zhang2025doors}.
Their articulated assets typically rely on simple target-position control,
equivalent to adding springs, together with global friction and damping parameters; this is executable but has limited expressiveness for structured joint dynamics. 
\methodname uses the simulator as the execution substrate, but
adds an interpretable and expressive field representation that can be generated,
edited, simulated, and optimized.

\paragraph{Joint-level interaction effects}
Measurements of everyday doors and drawers show diverse force trajectories with
low-dimensional structure, suggesting the potential for compact approximations
\citep{jain2010complex}. In engineering mechanics, reduced or equivalent force
models are also a central tool for representing complex joint interfaces without
simulating every contact detail \citep{brake2018jointedstructures}. Thus, rather than
reconstructing internal mechanisms, \methodname models their equivalent
joint-level effects.
Industrial hinge and lid-support catalogs organize such effects into recurring
families, including detent, bi-stable, torque, counterbalance, soft-down, and
lift-assist mechanisms \citep{southco2021hinges,sugatsune_lid_supports}. They provide engineering grounding for our templates, which compose
behavioral priors into force, friction, and damping fields rather than
identifying the exact hidden mechanism.

\section{Method}
\label{sec:method}
\begin{figure}[t]
  \centering
  \includegraphics[width=1.0\linewidth]{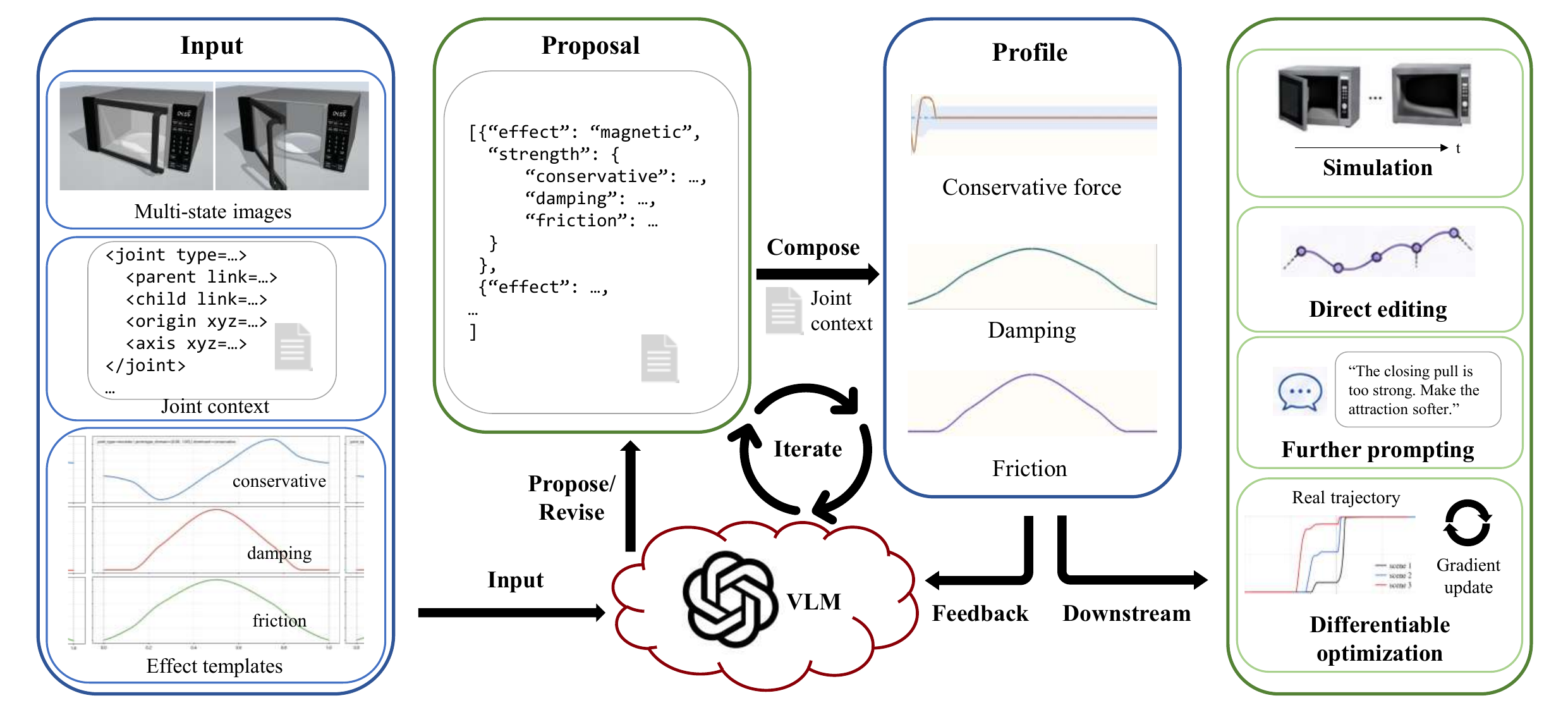}
  \caption{Overview of \methodname. A vision-language model proposes and iteratively refines structured joint effects from multimodal inputs, which are composed into a three-channel dynamics profile enabling simulation, editing, and optimization.}
  \label{fig:pipeline}
\end{figure}

Given joint context and visual observations, \methodname infers a structured dynamical representation that captures realistic interaction behavior. Fig.~\ref{fig:pipeline} gives an overview of our pipeline.

\subsection{Problem Formulation}

The input to \methodname is an articulated asset with a selected target joint. 
The asset provides a visual model, a kinematic hierarchy, and a finite joint range.
From this asset, we extract joint
context, including joint type, parent and child bodies, joint axis, joint range,
equivalent inertia or mass, and gravitational effects along the
degree of freedom (DOF), which can be directly obtained from standard simulation or modeling tools like MuJoCo.

The output represents the internal dynamical structure of that joint. It is internal in the
sense that it represents object-specific effects such as hinge drag, latching,
or return tendencies, not external forces such as gravity. These effects are expressed as generalized forces or coefficients
along a single DOF. 
Joint-limit behavior is modeled separately, as it is conventionally defined at the kinematic level as boundary conditions.

\subsection{Joint-Level Dynamical Representation}

Let \(q\) denote the generalized coordinate of the target joint, with finite
range \([q_{\min}, q_{\max}]\). We use a normalized coordinate
\[
  s = \frac{q - q_{\min}}{q_{\max} - q_{\min}}, \quad s \in [0,1].
\]
\methodname represents the internal dynamical structure of a joint with three
position-dependent fields:
\[
  F_{\mathrm{cons}}(s), \quad
  F_{\mathrm{fric,max}}(s), \quad
  C_{\mathrm{damp}}(s).
\]
The conservative channel \(F_{\mathrm{cons}}\) models position-dependent
generalized forces or torques, such as spring return, magnetic-like attraction,
detent wells, snap-through barriers, and bistable tendencies. The dry-friction
channel \(F_{\mathrm{fric,max}}\) models the maximum local friction magnitude. The
damping channel \(C_{\mathrm{damp}}\) models a velocity-dependent resistive
coefficient, producing a damping force proportional to \(-C_{\mathrm{damp}}(s)
\dot q\).

Each channel is composed from local effect components. A component \(j\) is
defined by an active interval \([a_j,b_j]\subseteq[0,1]\), a set of local
control-point positions \(x_{j,k}\in[0,1]\), and values \(y_{j,k}\). For
\(s\in[a_j,b_j]\), the local coordinate is
\[
  u = \frac{s-a_j}{b_j-a_j}.
\]
The component value is obtained via shape-constrained piecewise cubic interpolation (PCHIP), producing smooth curves over local control points while preserving physically plausible profiles.
Outside \([a_j,b_j]\), the component contributes zero. The
channel-level field is obtained by summing or
taking a maximum over active components. This representation is low-dimensional,
local, interpretable, and editable: changing a few control-point values or an
active interval changes a specific portion of the joint behavior without
requiring a global curve redesign.

\subsection{Effect Template Library}

\methodname uses a composable library of \emph{effect-templates}. The term effect is deliberate:
a template is not a claim that a particular hidden mechanical mechanism exists
inside the object. Instead, it specifies an equivalent joint-level behavior that
may be produced by many physical designs. For example, an endpoint attraction
template may approximate a magnetic catch or a spring-loaded return mechanism.
This mechanism-agnostic design makes our effect vocabulary more transferable.

Each template contains semantic metadata and prototype curves for the
three channels. The metadata includes the effect name, a short description,
and typical placement priors. The prototype
curves define the shape of the conservative-force, friction, and damping
contribution in local normalized coordinates. Instantiating a template means placing it on a selected interval of the target joint and scaling each channel.

The current library covers a range of common joint behaviors, including constant friction, constant damping, constant
directional conservative force, endpoint magnetic-like return, spring-like
return, snap detent, internal detent, and bistable effects. 
A summary of representative templates and their corresponding behaviors is provided in App.~\ref{sec:prompt}.
Higher-level interaction behaviors can be decomposed into these primitive effect templates.
For example, a free-stop experience is typically represented by strong constant
friction, possibly accompanied by damping.

\subsection{VLM-guided Effect Proposal}

The VLM stage predicts structured effect proposals.
Its inputs are the joint context, rendered images at multiple joint
states, and effect-template previews.
The prompt asks the VLM to reason about the
felt behavior a user would experience while moving the joint across its range.
The VLM is explicitly instructed not to output final control points or force curves. Instead, it outputs JSON containing effect proposals.
A summary of the prompt design is provided in App.~\ref{sec:prompt}, with the full prompt and schema in the supplementary material.

Each effect proposal specifies an effect name from the library, an active interval
$($\texttt{start\_ratio}, \texttt{end\_ratio}$)$,
and qualitative strength labels for the conservative, friction, and damping
channels. The proposal also includes descriptions like reasons and evidence which
are useful for inspection.

The output separates semantic choice from numerical realization. This division
reduces the burden on the VLM: it need not synthesize physically scaled curves
directly. It only chooses which equivalent effects should exist, where they act,
and how strong they are qualitatively.

\subsection{Compilation to Dynamics}

The compiler maps VLM proposals and joint context into a concrete
multi-component field. It converts normalized
intervals to real joint intervals and scales each channel into physical units.

The compiler is semi-deterministic.
Qualitative strength labels are mapped to numeric multipliers by sampling
within predefined bands, while the sample is generated from a stable key containing some meta-info.
This preserves repeatability while avoiding one-to-one mapping from strength labels to a deterministic value.

To set physical scale, the compiler computes reference magnitudes from joint
context. For conservative force and friction, it uses the larger of an estimated gravity
scale and an inertial scale based on equivalent inertia or mass, joint range,
and a reference travel time. The damping reference is chosen so that multiplying
it by a reference speed yields the reference force. Each template
channel is then scaled so that its prototype peak reaches the target channel
reference multiplied by the sampled multiplier.
The result is a \texttt{composed.json} file containing the final three-channel PCHIP field model.
This file is the source for downstream plotting and simulation.

\subsection{Diagnostic Feedback and Iterative Proposal}
After a proposal is compiled, JODA renders a diagnostic visualization of the resulting joint-level dynamics and feeds it back to the VLM. 
This visualization (e.g., Fig.~\ref{fig:lighter_profiles_and_interaction}c) illustrates the composed effects across the joint DOF and provides a structured signal for assessing whether the generated dynamics match the intended behavior.
We provide detailed interpretations of the diagnostic visualization in Sec.~\ref{sec:case_analysis} alongside concrete examples.
Based on this feedback, the VLM can iteratively revise the proposal by adjusting the effect types, active intervals, strength labels, or channel-specific refinement factors. 
This closed-loop iteration enables the system to progressively align the generated dynamics with the desired interaction behavior.

\subsection{Interactive and Differentiable Refinement}

Refinement after VLM generation pipeline may operate on 
variables include control-point values, channel strengths,
active interval anchors, and joint-limit damping parameters.
\methodname supports three refinement modes.

\textbf{Direct editing.}
An experienced user can directly modify the 
values
mentioned above.

\textbf{Human-supervised VLM refinement.}
A user can describe a problem in language, such as ``the final closing pull is
too strong'', or ``the middle range should hold better''. The VLM then revises its proposals while preserving the
overall intended motion when appropriate. The user can also specify the desired
effects over the full DOF, guiding the model to choose among multiple
plausible effects.

\textbf{Differentiable simulation optimization.}
VLMs provide strong semantic priors for motion, which can be further refined when higher physical accuracy is required. While manual adjustment is possible, it can be time-consuming, especially under quantitative constraints. Differentiable simulation offers a general mechanism for optimizing joint profiles under diverse supervision signals. In particular, leveraging real-world interaction trajectories as supervision targets enables direct alignment with observed motion, eliminating the need for manual tuning.

\section{Experiments}
\label{sec:experiments}

\subsection{Experimental setup and baselines}
\label{sec:exp_detail}

We evaluate \methodname on nontrivial object categories in PartNet-Mobility
\citep{xiang2020sapien} and Lightwheel Sim-Ready Assets
\citep{lightwheel_simready_2025}, processing one target joint per run.
We use two API models, \texttt{gemini-3.1-pro-preview} with the default reasoning
setting and \texttt{gpt-5.4} with high reasoning; we did not observe a clear
difference in task-level capability between them.
Our differentiable dynamics is implemented on top of the MuJoCo/MJX backend \citep{todorov2012mujoco,deepmind2024mjx}.
All experiments are run on a
laptop with an AMD Ryzen 9 8945HX CPU (16 cores/32 threads) and 32GB RAM.
Implementation details are provided in App.~\ref{app:composed_dynamics}.

Our first baseline removes the generated internal dynamical structure and
simulates the same articulated asset only with simple constant friction or damping. The second kind of baseline adds a linear return
force over the relevant joint interval, i.e., a spring-style model commonly used
in simulation-trained manipulation setups \citep{zakka2023robopianist,zhang2025doors}.

\begin{figure}[t]
    \centering
    \footnotesize
    \begin{minipage}[c]{0.12\linewidth}
        \centering
        \begin{overpic}[width=\linewidth]{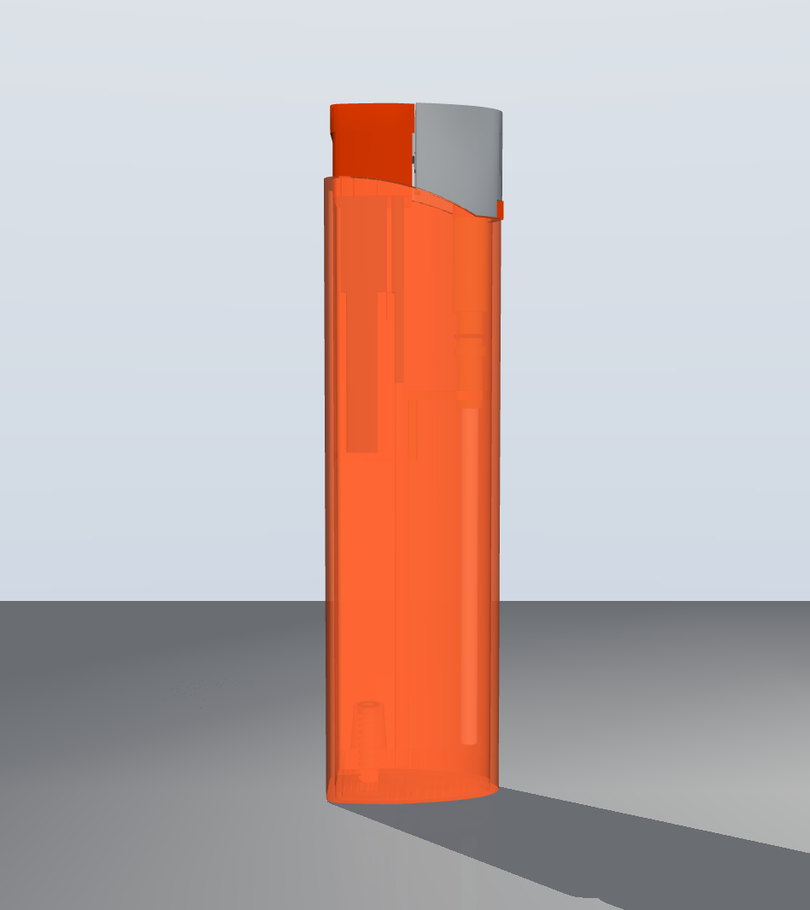}
        \put(0,50){\textbf{(a)}} 
        \put(0,35){\textbf{Sim}} 
        \end{overpic}        
        \begin{overpic}[width=\linewidth]{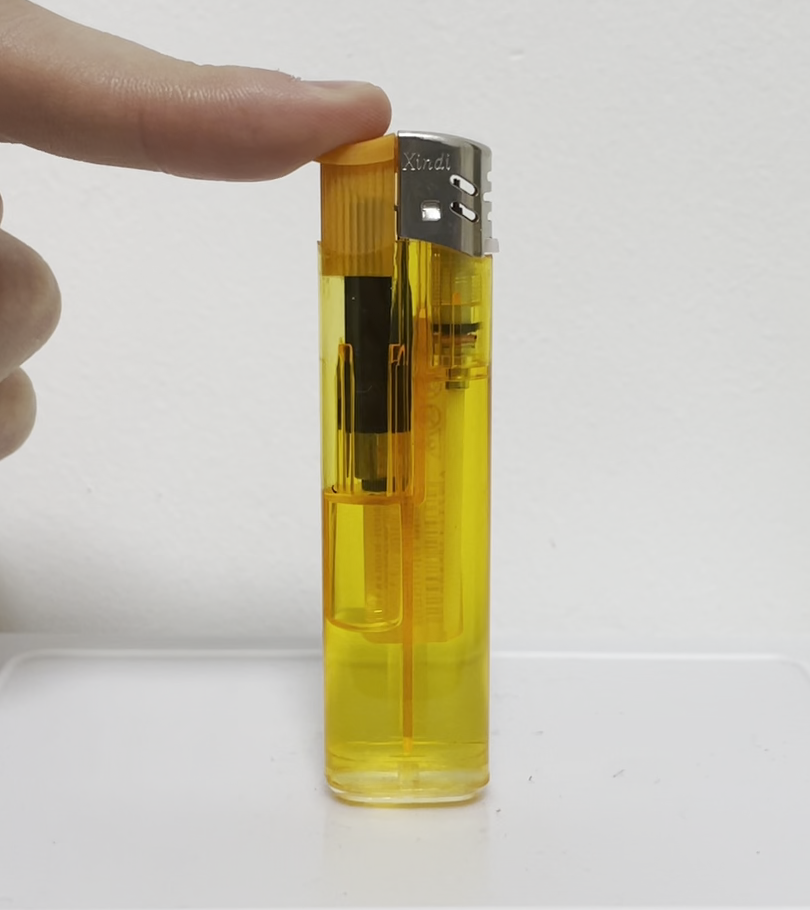}
        \put(0,35){\textbf{(b)}} 
        \put(0,20){\textbf{Real}}
        \end{overpic}
    \end{minipage}
    \hfill
    \begin{minipage}[c]{0.42\linewidth}
        \centering
        \begin{overpic}[width=\linewidth, trim={10pt 5pt 8pt 5pt}, clip]{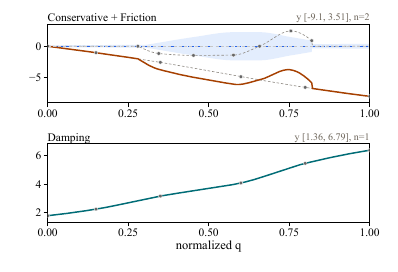}
            \put(30,35){\textbf{(c) Generated Profiles}}
        \end{overpic}
    \end{minipage}
    \hspace{2pt}
    \begin{minipage}[c]{0.44\linewidth}
        \centering
        \textbf{(d) Interaction force and resulting trajectory $q$}
        \includegraphics[width=\linewidth, trim={10pt 5pt 10pt 2pt}, clip]{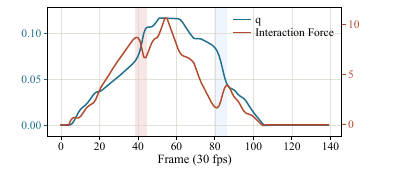}
    \end{minipage}
    
    \vspace{-6pt}
    \caption{\textbf{Lighter-button dynamics:} (a) simulated and (b) real-world visuals; (c) \methodname generated force and friction profiles; (d) Force and trajectory of the virtual interaction.} \label{fig:lighter_profiles_and_interaction}
\end{figure}

\subsection{Case analysis}
\label{sec:case_analysis}
For clearer comparisons of the dynamic motion, we refer the reader to the supplementary video \emph{Results (1/3): Baseline Comparisons}.
We visualize the generated dynamics using a diagnostic profile (e.g., Fig.~\ref{fig:lighter_profiles_and_interaction}c). 
This profile summarizes the composed joint-level dynamics across the DOF and is also used by VLM to determine whether further refinement is needed.
In this visualization, the red curve represents the conservative force, while the blue dashed line indicates the gravity-induced equilibrium. The shaded band reflects the friction range, within which small force imbalances may not lead to motion. When the red curve lies above the blue dashed line, the net effect drives the joint in the positive direction; when it lies below, it drives motion in the negative direction.
This visualization provides an interpretable link between generated dynamics and the resulting behavior.

\paragraph{Lighter button}

We first examine the push button of a cigarette lighter, whose target joint is a
short-travel prismatic DOF, as shown in
Fig.~\ref{fig:lighter_profiles_and_interaction}a. Fig.~\ref{fig:lighter_profiles_and_interaction}c
shows the compiled profile. The dominant effect is a full-range
spring-like return toward the unpressed state, while an additional internal
snap-like structure creates a non-monotone modification of the restoring force
in the middle part of the travel.

To probe the resulting interaction behavior, we use a virtual-hand interaction.
This interaction mode is designed to mimic a human hand: it moves a bounded, rate-limited control target,
and the simulator generates the corresponding interaction force needed to push
or release the button to reach that target using PD control. Fig.~\ref{fig:lighter_profiles_and_interaction}d plots the resulting
joint trajectory and interaction force over time.

\begin{figure}[t]
    \centering
    \begin{minipage}[t]{0.34\linewidth}
        \centering
        \begin{overpic}[width=1.1\linewidth]{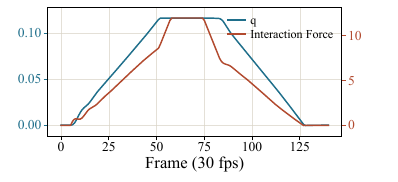}
            \put(14,35){{(a)}}
        \end{overpic}
        \begin{overpic}[width=\linewidth,clip,trim={15 0 0 0}]{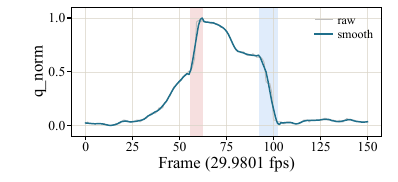}
        \put(12,25){{(b)}}
        \end{overpic}
        \\
        \vspace{-6pt}
        \captionof{figure}{\footnotesize
        Lighter button trajectories: (a) spring baseline; (b) real-world
        data (\texttt{q\_norm}: raw, \texttt{smooth}: smoothed).
        }
        \label{fig:lighter_baseline_and_real}
    \end{minipage}
    \hfill
    \begin{minipage}[t]{0.64\linewidth}
            \centering
        \begin{minipage}[c]{0.36\linewidth} \centering 
            \begin{overpic}[width=\linewidth,clip,trim={120 50 120 60}]{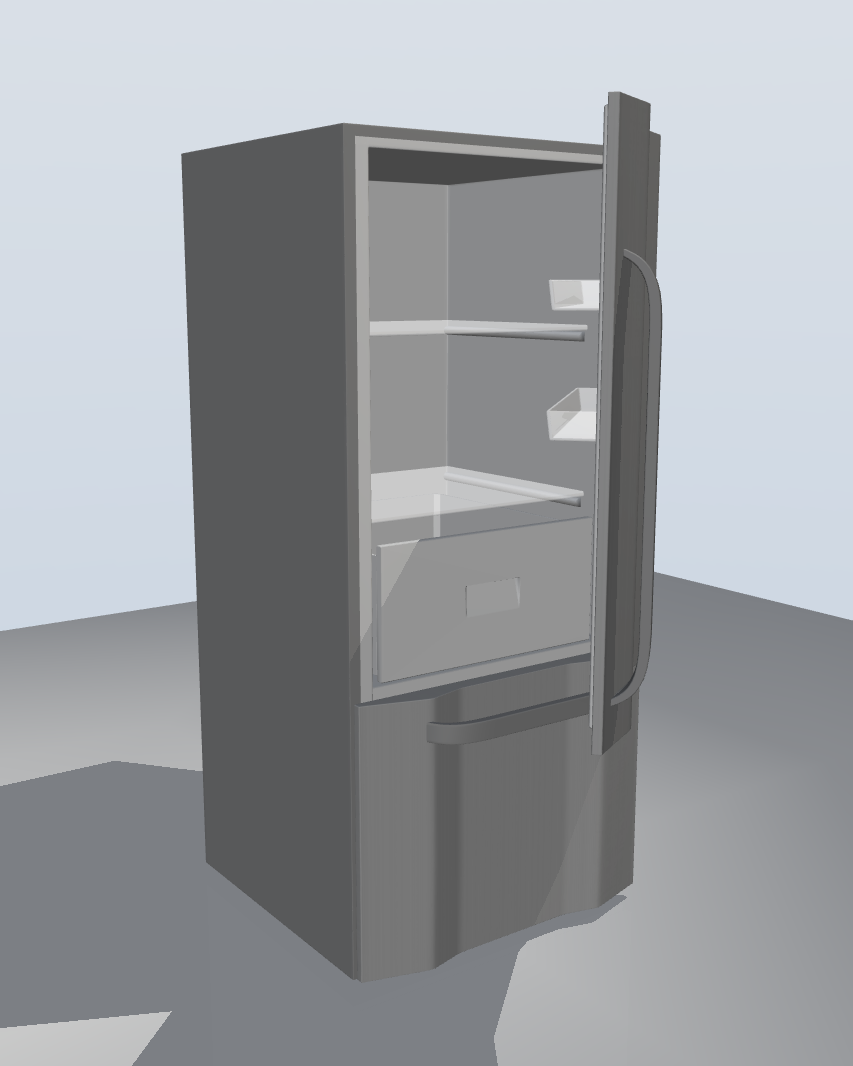} 
            \put(1,92){{(a)}} 
            \end{overpic} \vspace{6pt} 
        \end{minipage} \hfill 
        \begin{minipage}[c]{0.62\linewidth} \vspace{-6pt} 
            \begin{overpic}[width=0.95 \linewidth,clip,trim={0 0 0 0}]{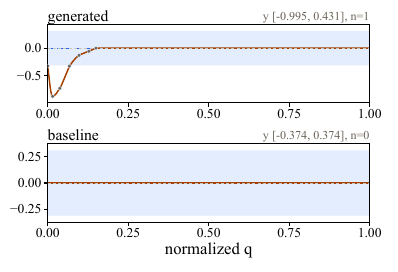}
            \put(2,60){{(b)}} 
            \put(2,32){{(c)}} 
            \end{overpic}             
            \begin{overpic}[width=0.9\linewidth,clip,trim={0 0 0 2}]{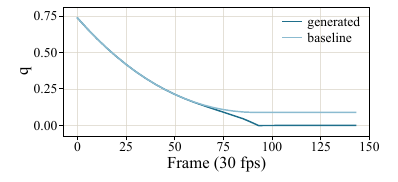} \put(2,45){{(d)}} 
            \end{overpic} 
            
        \end{minipage} \vspace{-6pt}
        \captionof{figure}{\footnotesize
        Refrigerator door: (a) visuals; (b, c) profiles; (d) trajectories.
        }
        \label{fig:refrigerator_combined}
    \end{minipage}
\end{figure}

The characteristic behavior of this button is that the internal mechanism
causes the effective rebound force to vary non-monotonically over the whole
DOF. As a result, both pressing and releasing can
produce short intervals of sudden high-speed motion, which is consistent with
the familiar push-through and pop-back feel of a lighter button. The shaded
regions in Fig.~\ref{fig:lighter_profiles_and_interaction}d mark these sudden high-speed
motion intervals.

We compare this result with a simple-spring baseline that uses only a broad
return spring, without the generated snap-like internal structure. The baseline
is evaluated with the same virtual-hand interaction controller. As shown in
Fig.~\ref{fig:lighter_baseline_and_real}a, the simple spring restores the
button but misses the distinctive push-through and pop-back transients.

We also use a real lighter, shown in Fig.~\ref{fig:lighter_profiles_and_interaction}b, to
verify that the button motion under human pressing exhibits a similar
push-through and pop-back pattern.
As shown in Fig.~\ref{fig:lighter_baseline_and_real}b, the simulated trajectory and
the real trajectory share the same qualitative behavior. The shaded regions indicate the two corresponding
sudden high-speed motion intervals.

\paragraph{Refrigerator door}
We evaluate a refrigerator door (Fig.~\ref{fig:refrigerator_combined}a). 
Our model captures a magnetic-seal effect that assists closing near the shut configuration, enabling the door to automatically close when nearly closed. 
In contrast, the baseline lacks such a structure and fails to close under the same condition. 
Fig.~\ref{fig:refrigerator_combined}d compares the resulting trajectories, while Fig.~\ref{fig:refrigerator_combined}b,c show the generated and baseline profiles, respectively.
\begin{figure}[t]
    \centering
    \footnotesize
    \begin{minipage}{0.32\linewidth}
        \centering
        \includegraphics[width=0.8\linewidth,clip,trim={30 140 30 140}]{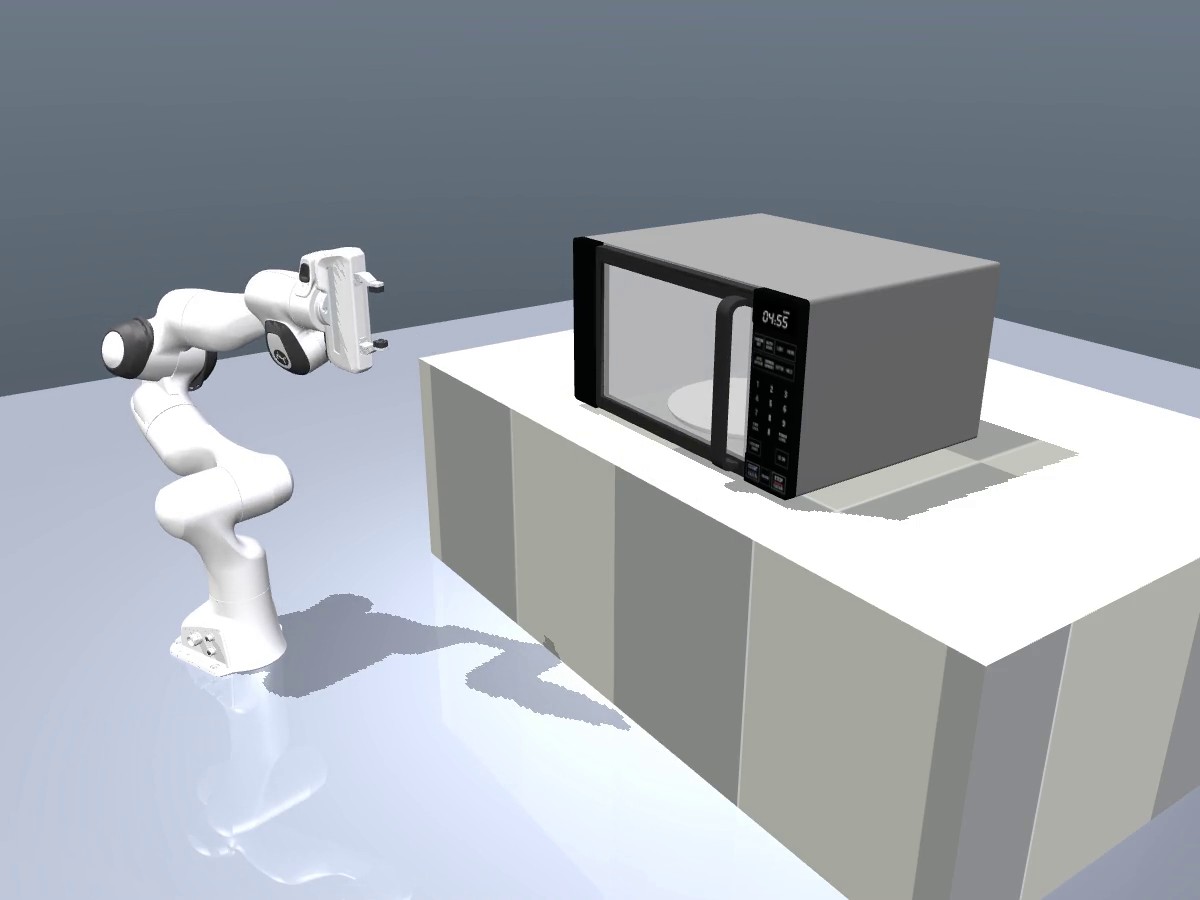}
        \vspace{2pt}
        \\
        \begin{overpic}[width=\linewidth,clip,trim={5 1 2 60}]{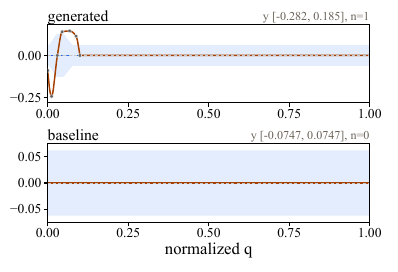}
            \put(3,86){(a) Baseline: unrealistic closing}
        \end{overpic}
        \vspace{-12pt}
        \\
        \includegraphics[width=\linewidth,clip,trim={5 1 2 1}]{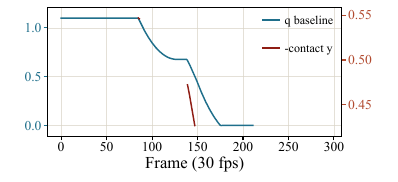}
    \end{minipage}
    \begin{minipage}{0.32\linewidth}
        \centering
        \includegraphics[width=0.8\linewidth,clip,trim={30 140 30 140}]{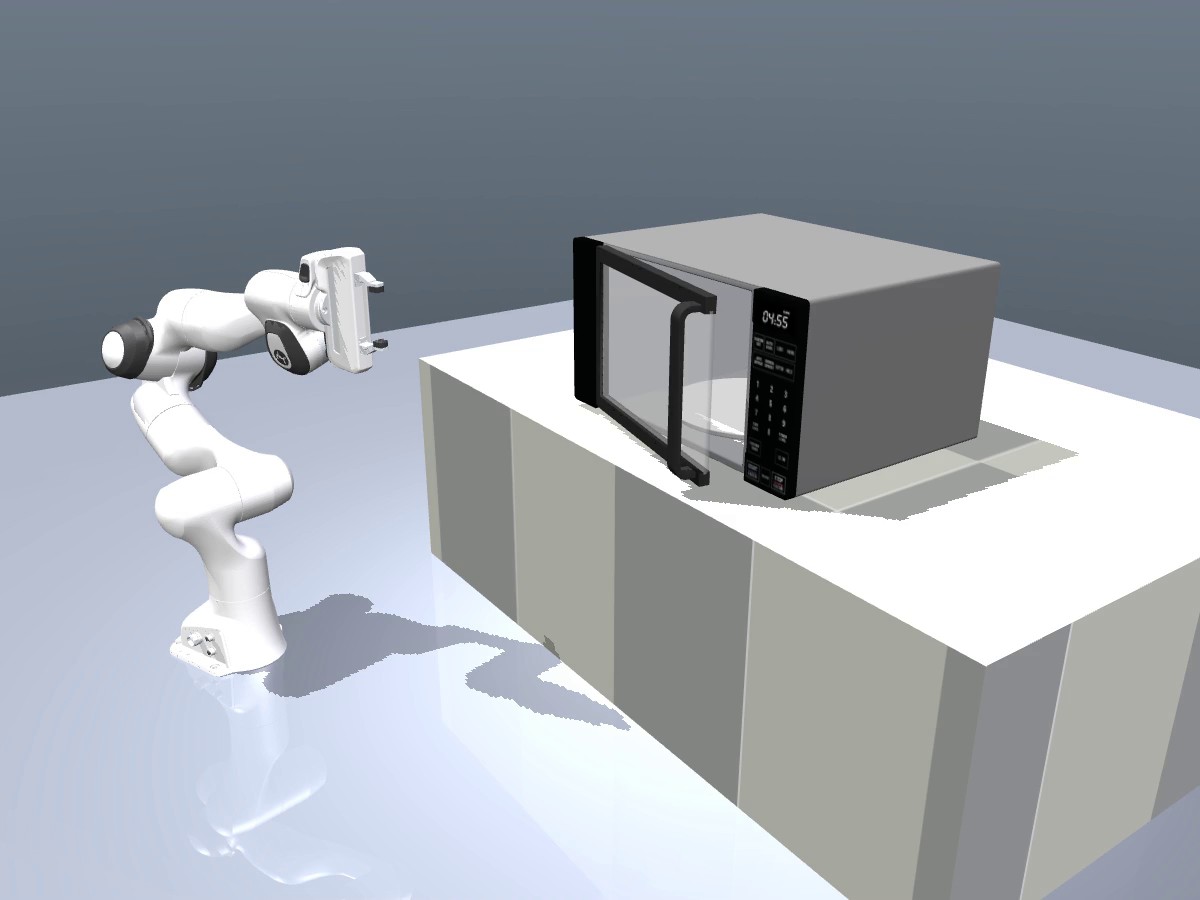}
        \vspace{2pt}
        \\
        \begin{overpic}[width=\linewidth,clip,trim={5 70 2 2}]{profile_overlay_generated_baseline_best.pdf}
            \put(3,80){(b) JODA: realistic rebound.}
            \put(40,-2){\tiny normalized q}
        \end{overpic}\vspace{5pt}
        \\
        \includegraphics[width=\linewidth,clip,trim={5 1 2 1}]{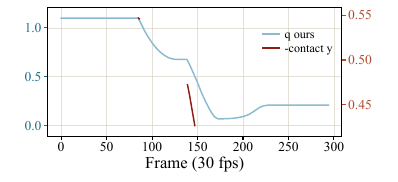}
    \end{minipage}
    \begin{minipage}{0.32\linewidth}
        \centering
        \includegraphics[width=0.8\linewidth,clip,trim={30 140 30 140}]{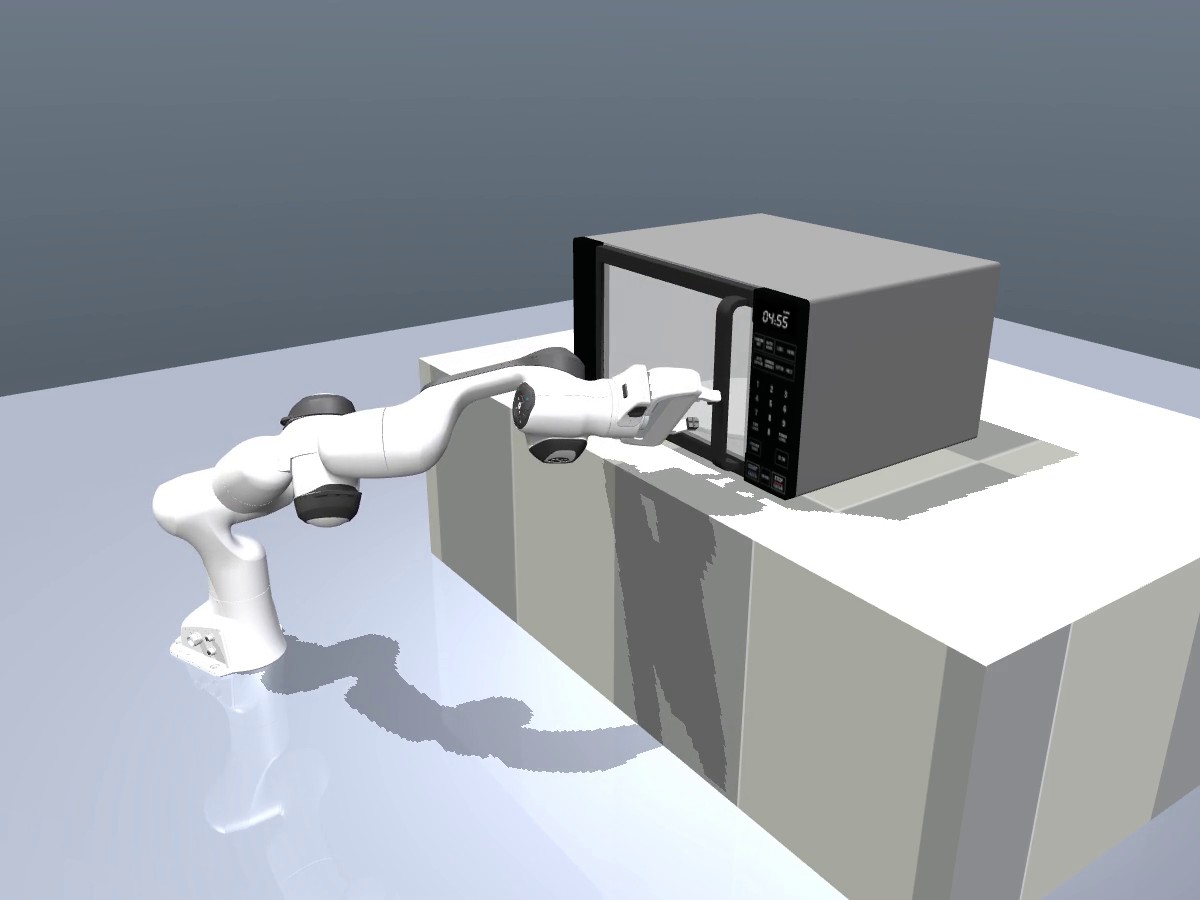}
        \vspace{2pt}
        \\
        \begin{overpic}[width=\linewidth,clip,trim={5 70 2 2}]{profile_overlay_generated_baseline_best.pdf}
            \put(3,80){(c) JODA: Closing requires an extra push.}
            \put(40,-2){\tiny normalized q}
        \end{overpic}\vspace{5pt}
        \\
        \includegraphics[width=\linewidth,clip,trim={5 1 2 1}]{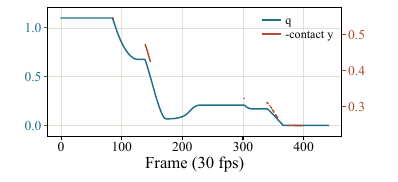}
    \end{minipage}
    \hfill
    \caption{\footnotesize Microwave door dynamics (visuals, profiles, and trajectories).
(a) Constant-damping baseline: the door automatically closes under slight pushes, which is unrealistic.
(b) JODA with the same robot motion offers a rebound near closure, preventing latching.
(c) JODA with an additional push: the door successfully latches after an extra contact.}
    \label{fig:merged_microwave}
\end{figure}

\paragraph{Microwave door}
We evaluate the generated dynamics on a microwave door (Fig.~\ref{fig:merged_microwave}). 
To evaluate interaction, we simulate a robot manipulator executing predefined pushing trajectories. 
JODA generates a profile that captures a spring-loaded latch near the closed configuration, introducing an energy barrier that must be overcome to fully close.
We compare JODA with a constant drag baseline.
Under the same slight push, the baseline exhibits unrealistic door closing in a simple slowing-down motion (Fig.~\ref{fig:merged_microwave}a), while our model produces a partial door closure with a rebound behavior near the end (Fig.~\ref{fig:merged_microwave}b).
Our model requires an additional push to fully latch (Fig.~\ref{fig:merged_microwave}c), which is consistent with real-world interactions. 
These results highlight the importance of physically faithful dynamics for reducing sim-to-real discrepancies and enabling better real-world transfer.

\subsection{Quantitative results}

\begin{figure}[tbh]
    \centering
    \begin{subfigure}[t]{0.23\linewidth}
        \centering
        \includegraphics[width=\linewidth,height=0.10\textheight,keepaspectratio]{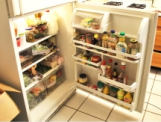}
        \caption{Jain fridge.}
    \end{subfigure}
    \hfill
    \begin{subfigure}[t]{0.23\linewidth}
        \centering
        \includegraphics[width=\linewidth,height=0.10\textheight,keepaspectratio]{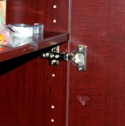}
        \caption{Jain cabinet.}
    \end{subfigure}
    \hfill
    \begin{subfigure}[t]{0.23\linewidth}
        \centering
        \includegraphics[width=\linewidth,height=0.10\textheight,keepaspectratio]{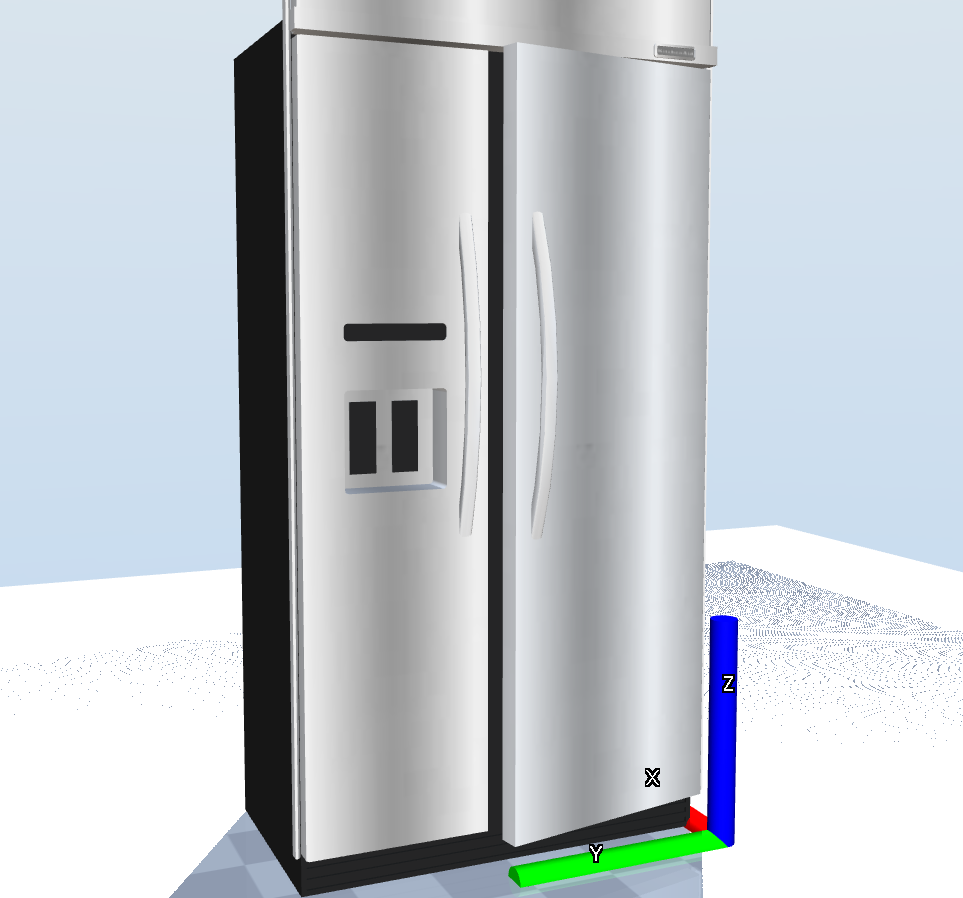}
        \caption{Our fridge render.}
    \end{subfigure}
    \hfill
    \begin{subfigure}[t]{0.23\linewidth}
        \centering
        \includegraphics[width=\linewidth,height=0.10\textheight,keepaspectratio]{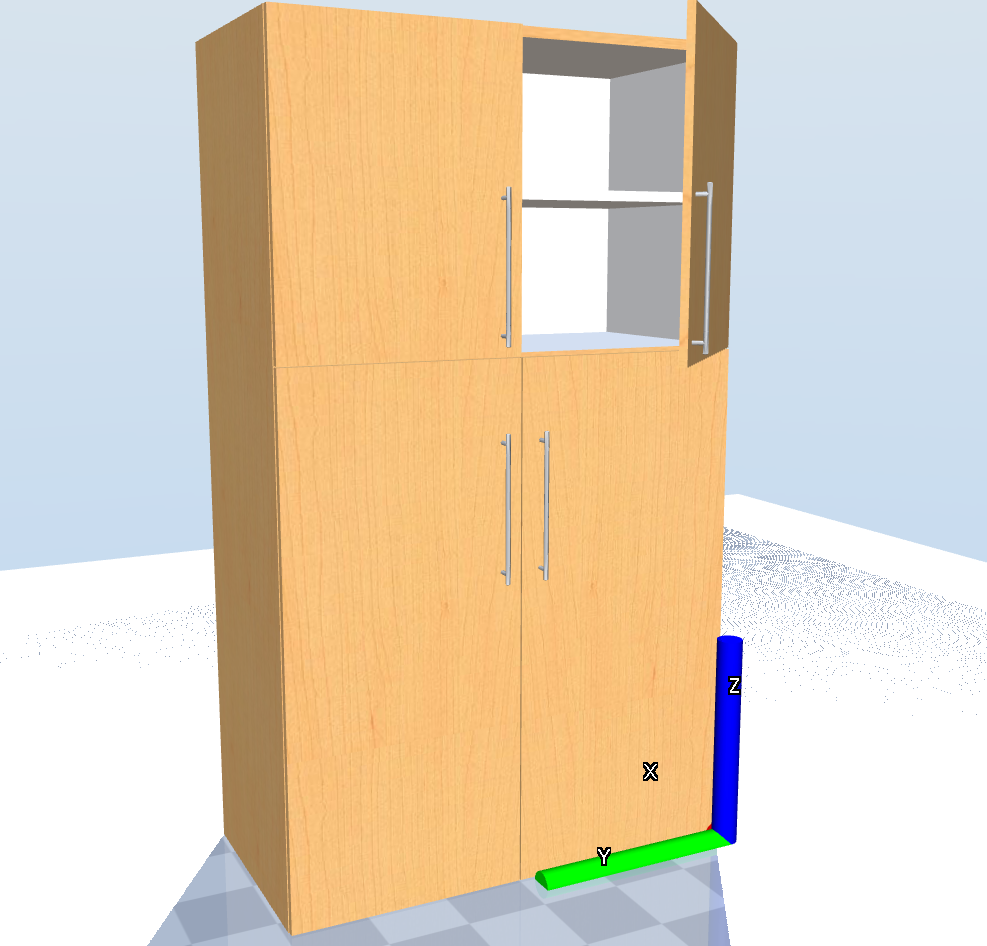}
        \caption{Our cabinet render.}
    \end{subfigure}
    \vspace{-6pt}
    \caption{Visual references for quantitative comparison.}
    \label{fig:jain2010_visual_comparison}
    \vspace{-10pt}
\end{figure}

\begin{figure}[tbh]
    \centering
    \begin{minipage}{0.79\linewidth}
    \begin{subfigure}[t]{0.5\linewidth}
        \centering
        \includegraphics[width=\linewidth, trim={11pt 7pt 2pt 3pt}, clip]{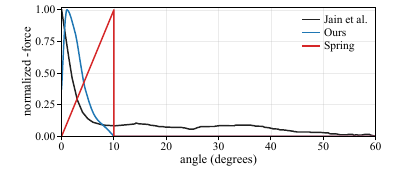}
    \vspace{-16pt}
        \caption{Refrigerator door.}
    \end{subfigure}
    \begin{subfigure}[t]{0.5\linewidth}
        \centering
        \includegraphics[width=\linewidth, trim={11pt 7pt 2pt 3pt}, clip]{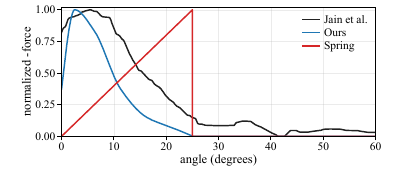}
    \vspace{-16pt}
        \caption{Kitchen cabinet door.}
    \end{subfigure}
    \end{minipage}
    \hfill
    \begin{minipage}{0.20\linewidth}
        \centering
        \phantomsubcaption
        \footnotesize
        \setlength{\tabcolsep}{1.5pt}
        \begin{tabular}{lcc}
            \toprule
            & Fridge & Cabinet \\
            \midrule
            Spring & 0.240 & 0.386 \\
            \methodname & \textbf{0.121} & \textbf{0.180} \\
            \bottomrule
        \end{tabular}
        \\\vspace{4pt}
        (c) Normalized force error w.r.t.~\citet{jain2010complex}.
        (Lower is better.)
    \end{minipage}
    \vspace{-8pt}
    \caption{Quasi-static opening force comparison against measurements from \citet{jain2010complex}.}
    \label{fig:jain2010_force_comparison}
\end{figure}

Our representation defines a joint-level dynamics profile, from which we can derive the quasi-static force that opens the door at each configuration. 
While this profile represents intrinsic joint behavior and is not directly measurable, prior work by \citet{jain2010complex} measured the corresponding quasi-static opening forces for everyday objects (e.g., refrigerator doors) in real-world settings.
We compare the forces derived from our generated profiles with these measured curves. Specifically, we normalize each curve to $[0,1]$, and evaluate \methodname against a linear spring baseline on a common $0$--$60^\circ$ grid. Fig.~\ref{fig:jain2010_visual_comparison} shows the
objects used for this comparison, and the first two images are real object photos from \citet{jain2010complex}. Fig.~\ref{fig:jain2010_force_comparison} compares the
quasi-static opening forces (positive values indicate external forces that increase the angle), and Fig.~\ref{fig:jain2010_force_comparison}c reports
the RMSE. \methodname more closely matches the measured curves, whereas a linear spring fails to capture the structured, nonlinear behavior, as it is constrained to produce zero force at equilibrium and linear force variation with displacement.
This demonstrates that expressive joint-level dynamics are necessary to reproduce real-world forces across configurations beyond simple models.

\begin{figure}[tbh]
    \centering
    \footnotesize
    \begin{minipage}[c]{0.16\linewidth}
        \centering
        \begin{overpic}[width=0.92\linewidth]{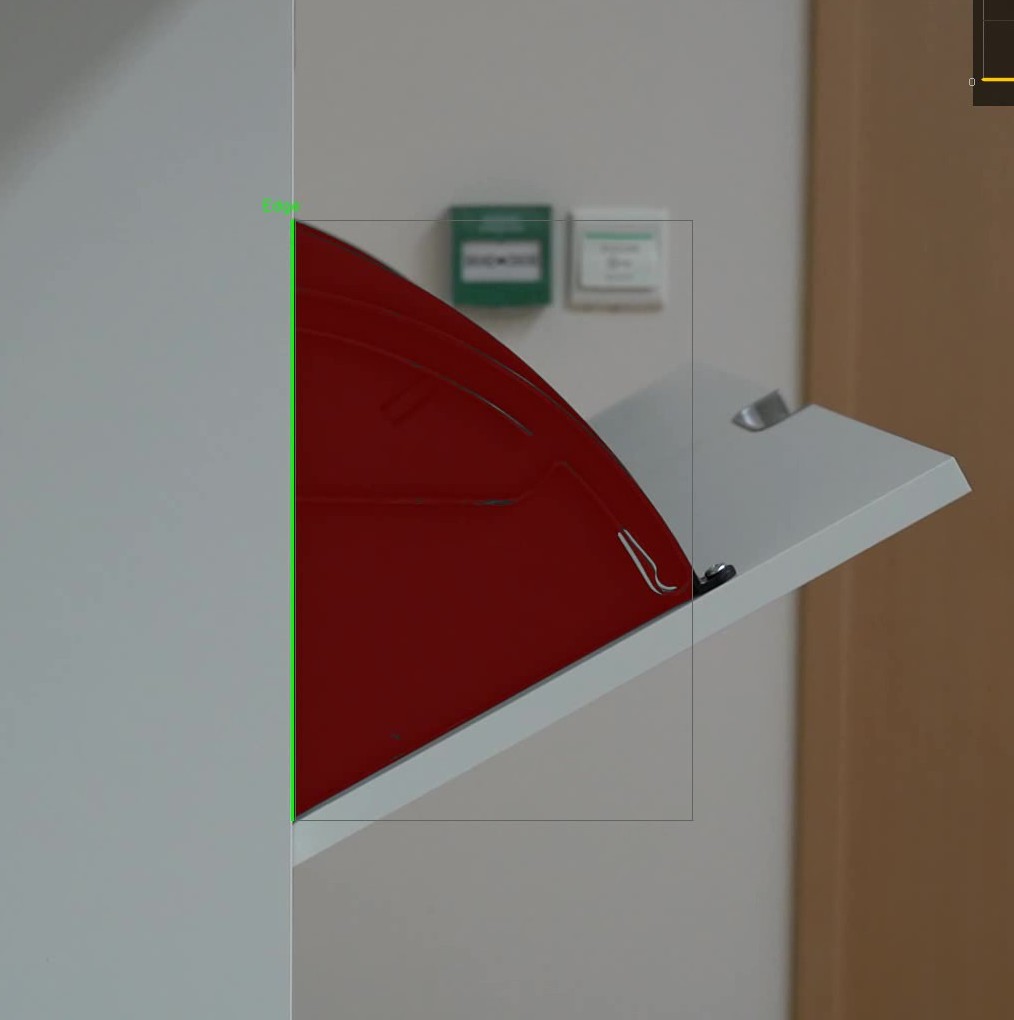}
            \put(2,88){\textbf{(a)}}
        \end{overpic}
        \\\vspace{2pt}
        \begin{overpic}[width=0.92\linewidth]{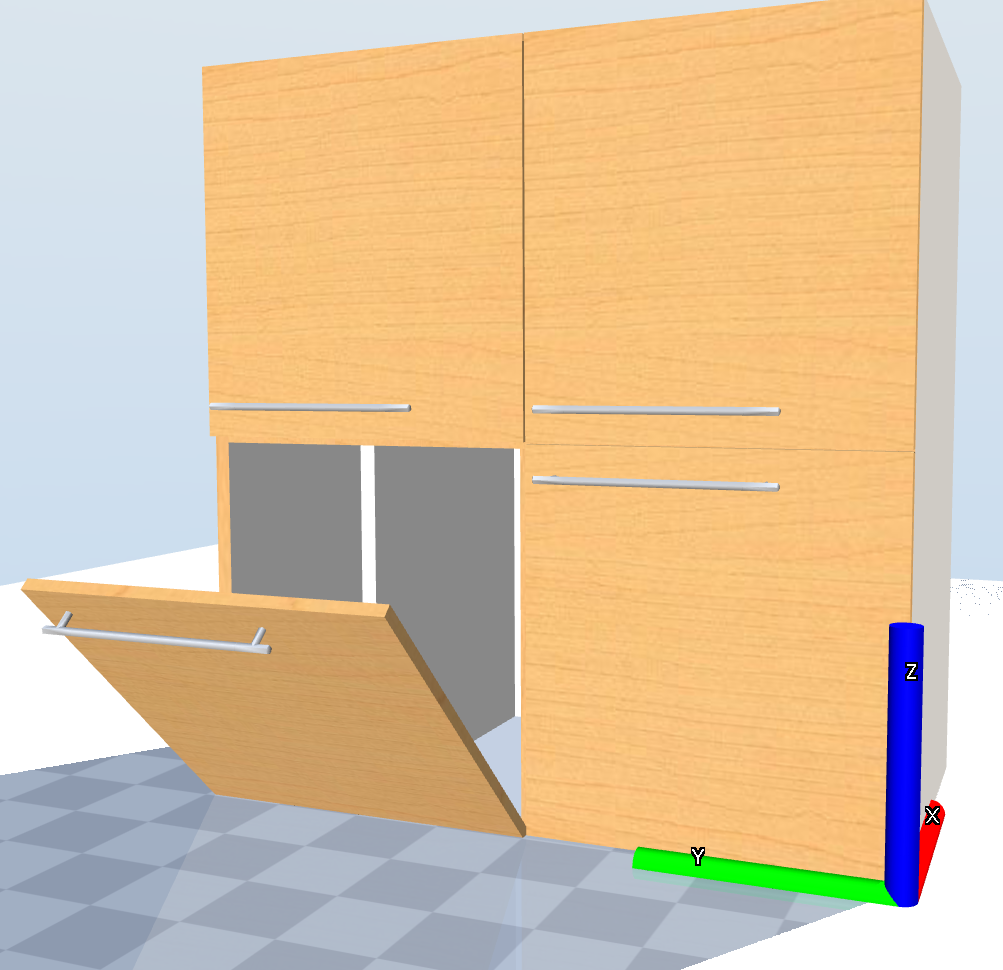}
            \put(2,86){\textbf{(b)}}
        \end{overpic}
    \end{minipage}
    \hfill
    \begin{minipage}[c]{0.39\linewidth}
        \centering
        \begin{overpic}[width=0.95\linewidth, trim={7pt 0pt 6pt 0pt}, clip]{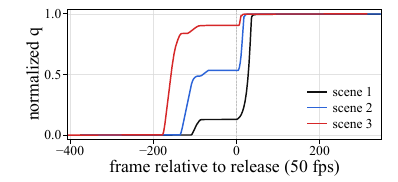}
            \put(0,44){\textbf{(c)}}
        \end{overpic}
        \begin{overpic}[width=\linewidth, trim={7pt 0pt 8pt 0pt}, clip]{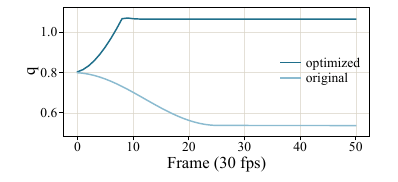}
            \put(0,42){\textbf{(d)}}
        \end{overpic}
    \end{minipage}
    \hfill
    \begin{minipage}[c]{0.42\linewidth}
        \centering
        \begin{overpic}[width=1.05\linewidth, trim={10pt 0pt 5pt 0pt}, clip]{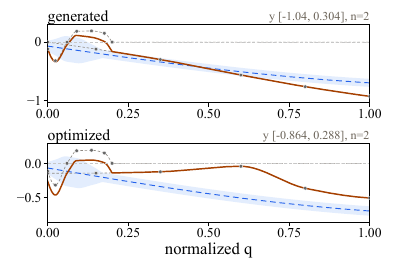}
            \put(2,68){\textbf{(e)}}
        \end{overpic}
    \end{minipage} 
    \vspace{-6pt}
    \caption{Differentiable refinement. (a,b) real and simulated visuals;
    (c) target trajectories; (d) release at q=0.8: original closes, optimized matches the observed opening behavior; (e) profiles.}
    \label{fig:oven_refinement_combined}
\end{figure}

We also measure computational cost. With a timestep of $dt=0.005$s, across all
runs with valid timing records, \methodname takes 0.062 ms per simulation step on
average, which is 1.37x the original MuJoCo cost, indicating only a modest
simulation overhead. For VLM proposal, across 252 proposal results, each
iteration uses on average 25.5K input tokens and produces 4.7K output tokens, where
output tokens include reasoning tokens.

\subsection{Refinement experiments and ablatioon studies}

\textbf{Refinements.} We experiment with a cabinet door that opens under gravity
(Fig.~\ref{fig:oven_refinement_combined}b). In
Fig.~\ref{fig:oven_refinement_combined}e, the model-generated result provides
too much support near the high-\(q\) end, preventing the door from opening
naturally. We sample three trajectories from a real cabinet instance
(Fig.~\ref{fig:oven_refinement_combined}a;
Fig.~\ref{fig:oven_refinement_combined}c, natural release at frame 0).
The trajectory MSE loss decreases from $4.10\times10^{-2}$ to
$1.56\times10^{-3}$ after 9 Adam iterations, and the optimized profile in Fig.~\ref{fig:oven_refinement_combined}e
opens naturally while retaining part of the support effect, giving a more
reasonable cabinet profile. Simulated trajectories with the original generated profile as well as the optimized one are shown in Fig.~\ref{fig:oven_refinement_combined}d.

\begin{figure}[h]
    \centering \footnotesize
    \begin{subfigure}[t]{0.32\linewidth}
        \centering        \includegraphics[width=1.05\linewidth, trim={7pt 6pt 8pt 6pt}, clip]{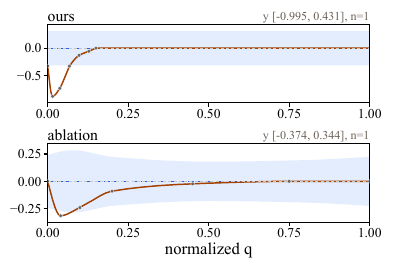}
        \vspace{-12pt}
        \caption{Refrigerator door.}
    \end{subfigure}
    \hfill
    \begin{subfigure}[t]{0.32\linewidth}
        \centering
        \includegraphics[width=1.05\linewidth, trim={8pt 6pt 8pt 6pt}, clip]{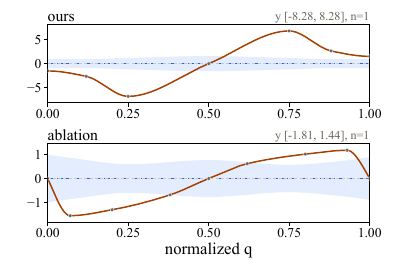}
        \vspace{-12pt}
        \caption{Switch.}
    \end{subfigure}
    \hfill
    \begin{subfigure}[t]{0.32\linewidth}
        \centering
        \includegraphics[width=1.05\linewidth, trim={7pt 6pt 8pt 6pt}, clip]{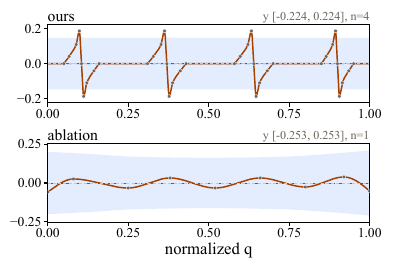}
        \vspace{-12pt}
        \caption{Selector knob.}
    \end{subfigure}
    \vspace{-8pt}
    \caption{Our results vs. No-template ablation results.}
    \label{fig:no_template_ablation}
\end{figure}

\begin{figure}[tbh]
    \centering
    \vspace{-12pt}
    \begin{minipage}[t]{0.6\linewidth}
        \centering
        \begin{subfigure}[t]{0.33\linewidth}
            \centering
            \includegraphics[width=\linewidth]{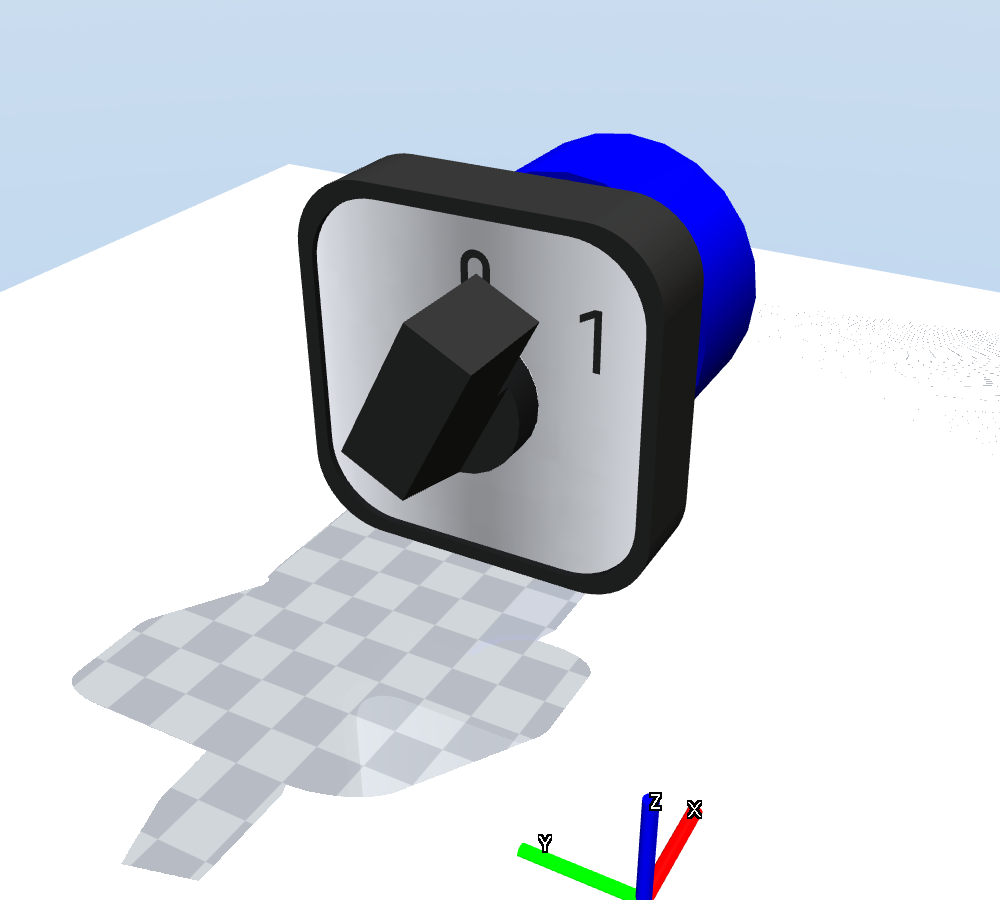}
            \caption{Switch}
        \end{subfigure}
        \hfill
        \begin{subfigure}[t]{0.31\linewidth}
            \centering
            \includegraphics[width=\linewidth]{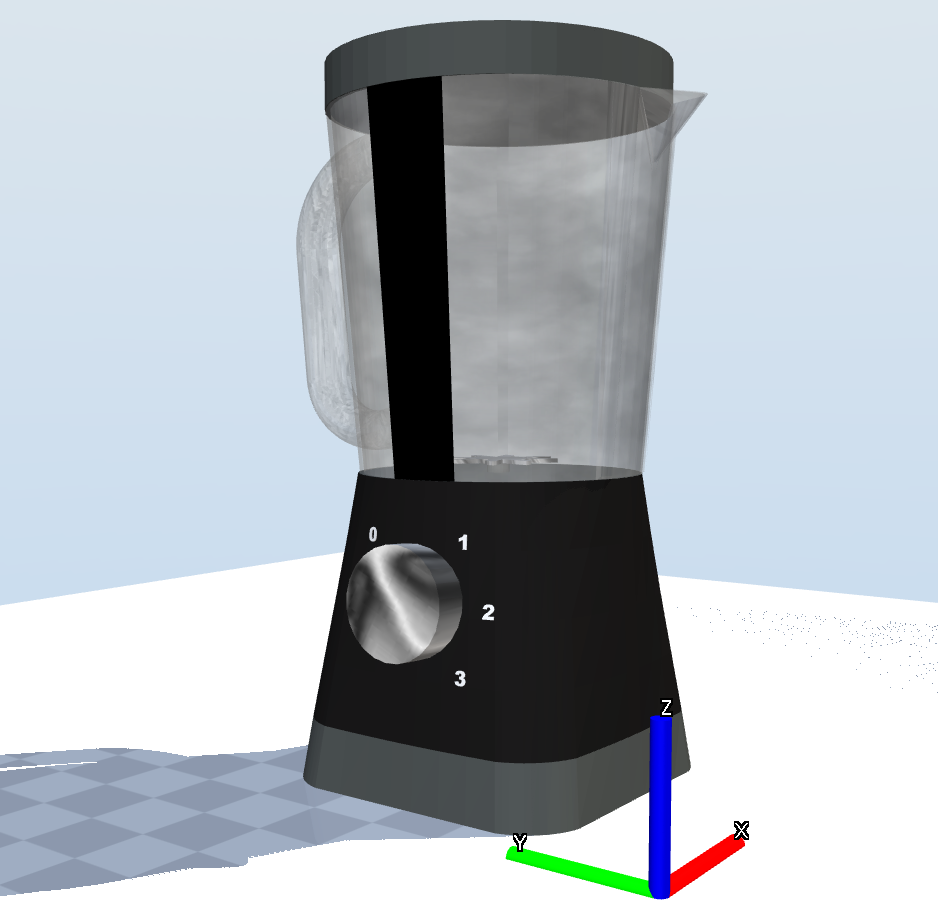}
            \caption{Selector}
        \end{subfigure}
        \hfill
        \begin{subfigure}[t]{0.33\linewidth}
            \centering
            \includegraphics[width=\linewidth]{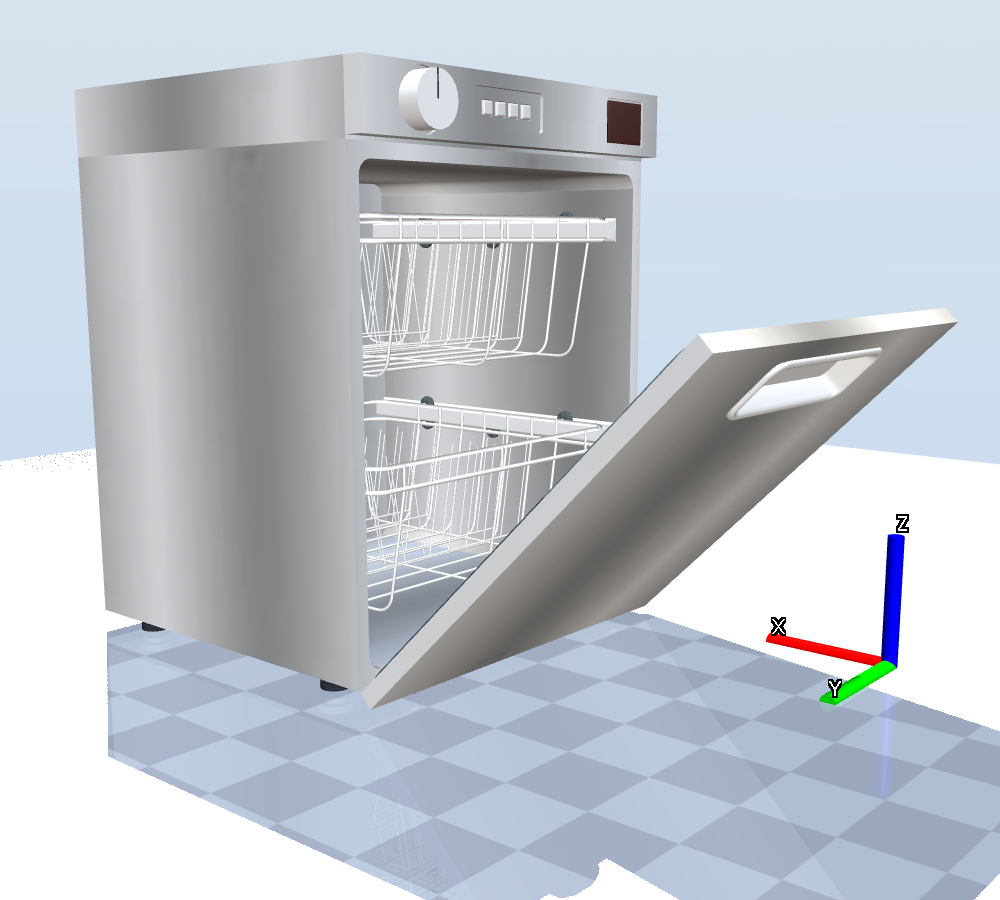}
            \caption{Dishwasher}
        \end{subfigure}
        \vspace{-6pt}
        \caption{Prompt images used in the ablation.}
        \label{fig:no_template_prompt_views}
    \end{minipage}
    \hfill
    \begin{minipage}[t]{0.36\linewidth}
        \centering
        \includegraphics[width=0.9\linewidth,clip,trim={10 6 10 2}]{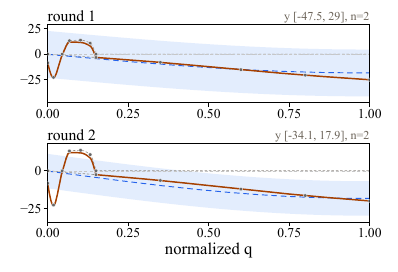}
        \vspace{-6pt}
        \captionof{figure}{Initial and iterated field profiles for a dishwasher
        door.}
        \label{fig:dishwasher_refinement_overlay}
    \end{minipage}
    \vspace{-18pt}
\end{figure}

\textbf{Ablations.} We first ablate the template library by directly asking the VLM to output
numerical control-point coordinates and then interpolating those points into
fields. This setting places a much heavier burden on the VLM's quantitative
reasoning and curve-generation ability: it often fails to choose appropriate
magnitudes, and without engineering priors it also introduces small but
behaviorally important defects. For example, in the refrigerator example
(Fig.~\ref{fig:no_template_ablation}a), the magnetic-attraction force is too
weak and acts over an overly broad range. In the switch example
(Fig.~\ref{fig:no_template_ablation}b; Fig.~\ref{fig:no_template_prompt_views}a), the conservative force is weak relative to
friction, and the force dropping to zero at the balance point makes the return
tendency too weak. In the selector-knob example
(Fig.~\ref{fig:no_template_ablation}c; Fig.~\ref{fig:no_template_prompt_views}b), the conservative force is again too weak
relative to friction. In all these examples, both \methodname and the ablations are evaluated without iterative rounds. 
Furthermore, removing the template library significantly degrades reproducibility.

The VLM iteration step is especially useful in scenes with gravity. In the
dishwasher-door example
(Fig.~\ref{fig:no_template_prompt_views}c),
Fig.~\ref{fig:dishwasher_refinement_overlay} shows that the round-1
near-closed attraction does not overcome friction, while the opening-end assist
is too large and creates an early closing tendency that may prevent the door
from fully opening. After iteration, the far-end assist decreases and the
near-closed assist increases, matching the intended effect. Across gravity
scenes, \methodname satisfies the initial effect specification in 2.2 proposal
rounds on average, showing that iteration is necessary.

\section{Conclusion}
\label{sec:conclusion}

We introduced \methodname, a framework for synthesizing
joint-level dynamics for articulated objects. Instead of modeling
hidden mechanisms, it represents interaction behavior as a
composition of conservative force, friction, and damping fields.
A VLM proposes structured effects from visual and joint context,
which are compiled into physically scaled fields and executed in
MuJoCo/MJX. This enables direct simulation and multiple refinement
modes, including manual editing, language-guided revision, and
differentiable optimization.

The current representation assumes that the dominant internal behavior can be
approximated along a single joint degree of freedom with three
position-dependent channels: conservative force, dry-friction magnitude, and
damping. This covers many common articulated interactions, but it does not yet
support more complex dynamical structures such as hidden-state effects,
nonlinear damping, complex or history-dependent friction, high-dimensional DOFs, and multi-joint
couplings.

Overall, \methodname turns joint dynamics into an authorable and
composable representation, extending articulated assets beyond geometry to executable interaction behavior. 
We view this as a step toward physically grounded, behavior-aware asset
generation for simulation, robotics, and interactive systems.
Future work includes expanding the effect vocabulary, modeling
multi-joint and mode-dependent dynamics, improving automatic
diagnostics, and evaluating perceptual alignment with human interaction expectations.

\section*{Acknowledgments}

This work was supported by the National Natural Science
Foundation of China (Project Number 62595771) and Ant Group Research Fund.

\bibliographystyle{plainnat}
\bibliography{references}

\clearpage\appendix

\section{Implementation of Composed Dynamics Fields}
\label{app:composed_dynamics}

\subsection{MuJoCo-compatible asset realization}

\methodname targets executable articulated assets. We use MuJoCo XML as the canonical
simulation representation.

The MuJoCo model serves as the source of joint context. We extract the
target joint's parent and child bodies, local and world joint axes, range,
equivalent generalized inertia or mass from the mass matrix, and an engineering
estimate of gravitational scale along the selected degree of freedom. This
context anchors the compiler's numeric scaling and helps the VLM interpret the
motion direction shown in rendered images.

\subsection{Differentiable MJX implementation of dynamical fields}

The composed field is executed in an MJX rollout. At each step, the PCHIP components are evaluated, and the
three channels are injected into the dynamics. The conservative channel is
written as an applied generalized force or torque. The damping channel updates
the selected degree of freedom's damping coefficient. The friction channel
updates the local dry-friction loss used by the solver. The joint-limit hint is
mapped to MuJoCo/MJX limit solver parameters, such as the damping ratio of the
selected joint limit model.

This implementation is designed to be differentiable-friendly. Field evaluation
uses JAX-compatible PCHIP interpolation, and the conservative and damping
channels are directly parameterized by control-point values that can be
optimized through MJX rollouts. Dry friction and joint-limit constraints are
less smooth and require local smoothing. The same
field representation provides a natural parameter space for gradient-based
refinement.

\section{VLM Prompt and Effect Proposal Specification}
\label{sec:prompt}

We design a structured prompting interface that enables the VLM to infer joint-level effects in a physically grounded and compositional form. Given joint context, multi-state rendered images, and effect-template previews, the model infers equivalent joint-level effects across the full joint range.

\subsection{Prompt Excerpt}
\label{app:prompt_excerpt}

We provide a shortened excerpt of the prompt used for VLM-based effect proposal in Fig.~\ref{fig:prompt_short}. The full prompt is provided in the supplementary material. The prompt casts the model as a structured mechanics analyst and enforces physically grounded reasoning and structured output. 

\begin{figure}[h]
\begin{promptbox}[Prompt Design]
\textbf{Role.}
You are a careful multimodal mechanics analyst working in a
high-precision pipeline. Your output will be used as structured input
to a downstream compiler. Prioritize correctness, consistency, and
physical plausibility over free-form creativity.
\newline\\
\textbf{Task.}
Given one articulated joint, infer the equivalent joint-level effects
that describe the felt behavior along the entire motion range. These
effects represent observable interaction behavior, not explicit claims
about hidden mechanisms.
\newline\\
\textbf{Inputs.}
\begin{itemize}
\item Joint context (type, axis, range, gravity relation);
\item multi-state rendered images along the joint coordinate;
\item template previews showing reference force, friction, and damping profiles.
\end{itemize}
\vspace{6pt}
\textbf{Output.}
Return a JSON object (detailed in App.~\ref{sec:prompt_json}) describing:
\begin{itemize}
\item a whole-motion description,
\item a gravity-importance judgment,
\item a joint-limit hint,
\item a set of effect proposals (type, interval, qualitative strength).
\end{itemize}
\vspace{6pt}
\textbf{Constraints.}
\begin{itemize}
    \item Use only predefined effect templates (summarized in Table~\ref{tab:prompt_template}).
    \item Represent behavior as equivalent joint-level effects.
    \item Reason over the full motion range.
    \item Prefer simple explanations when multiple effects are plausible.
\end{itemize}
\vspace{6pt}

\textbf{Iterative refinement.}
When provided with a rendered feedback result, either return
\texttt{complete} if the behavior matches the intent, or revise the
effect proposals to better match the desired motion.
The full prompt is provided in the supplementary material.
\end{promptbox}
\caption{Prompt design for multimodal mechanics analysis.}
\label{fig:prompt_short}
\end{figure}

\begin{table}[htbp]
\centering
\small
\caption{Predefined effect templates defining the joint-level behavior vocabulary.}
\label{tab:prompt_template}
\begin{tabular}{p{5.2cm} p{7.5cm}}
\toprule
\textbf{Template} & \textbf{Behavior description} \\
\midrule
constant\_friction\_hinge & Constant dry friction over the interval \\
constant\_damping\_hinge & Constant velocity-dependent damping \\
constant\_positive\_conservative\_hinge & Constant drive toward increasing $q$ \\
constant\_negative\_conservative\_hinge & Constant drive toward decreasing $q$ \\

detent\_internal & Local click-stop with interior equilibrium \\
bistable\_mechanism & Two stable regions separated by a barrier \\
bistable\_mechanism\_internal & Two interior stable points \\

magnetic\_return\_to\_low\_end & Increasing attraction toward low end \\
magnetic\_return\_to\_high\_end & Increasing attraction toward high end \\

spring\_return\_to\_low\_end & Spring-like return toward low end \\
spring\_return\_to\_high\_end & Spring-like return toward high end \\

spring\_loaded\_snap\_detent\_to\_low\_end & Snap-in latch near low end \\
spring\_loaded\_snap\_detent\_to\_high\_end & Snap-in latch near high end \\
\bottomrule
\end{tabular}
\end{table}

\subsection{Output JSON Schema (Compact)}
\label{sec:prompt_json}

The VLM outputs a structured JSON object describing joint-level
dynamical effects, which is summarized in List~\ref{lst:json}.
The full JSON schema is provided in the supplementary material.
This schema separates semantic effect specification from numerical
realization. Qualitative strength labels are later mapped to physically scaled parameters by the compiler.

\begin{lstlisting}[language=json, caption={\textbf{VLM Output JSON Schema.} The structured 
output separates semantic effect specification from numerical realization.}, 
label={lst:json}]
{
  "joint_summary": {
    "joint_name": str,
    "joint_type": "revolute | prismatic",
    "motion_description": str,
    "overall_confidence": float
  },

  "whole_motion_descrptn": str,
  "gravity_can_be_ignored": bool,

  "joint_limit_hint": {
    "selected_side": "low_end | high_end | none",
    "elasticity": "none | weak | medium | strong"
  },

  "effect_proposals": [
    {
      "effect_name": str,                // from template library
      "start_ratio": float,              // in [0, 1]
      "end_ratio": float,                // in [0, 1]

      "strength": {
        "conservative": "none | weak | medium | strong | dominant",
        "friction":     "none | weak | medium | strong | dominant",
        "damping":      "none | weak | medium | strong | dominant"
      },

      "refineFactor": {
        "conservative": float,
        "friction": float,
        "damping": float
      },

      "confidence": float,
      "reason": str
    }
  ]
}
\end{lstlisting}

\end{document}